# Computational Scenario-based Capability Planning

[†]Hussein A. Abbass, [‡]Axel Bender, [†]Helen Dam, [‡]Stephen Baker, [†]James Whitacre, and [†]Ruhul Sarker

[†]School of Information Technology, University of New South Wales at Australian Defence Force Academy, Northcott Drive, Canberra, Australia, 61 2 626 88054, {h.abbass,h.dam, j.whitacre}@adfa.edu.au

[‡]Defence, Science and Technology Organisation, Edinburgh, Adelaide, Australia, 61 8 825 96341, axel.bender@dsto.defence.gov.au

## ABSTRACT
Scenarios are pen-pictures of plausible futures, used for strategic planning. The aim of this investigation is to expand the horizon of scenario-based planning through computational models that are able to aid the analyst in the planning process. The investigation builds upon the advances of Information and Communication Technology (ICT) to create a novel, flexible and customizable computational capability-based planning methodology that is practical and theoretically sound. We will show how evolutionary computation, in particular evolutionary multi-objective optimization, can play a central role – both as an optimizer and as a source for innovation.

## Categories and Subject Descriptors
G.1.6 **Optimization**: Stochastic programming

## 1. Introduction
Capability planning is a form of strategic/long-term planning [19] and the term is normally used interchangeably with defense planning. It focuses on the development of future plans and does not address the issue of making specific future decisions [19][20], which is normally taken care of in the scope of *operations planning*. The Australian Defence Force (ADF) combines both types of planning in a unified framework so that the planning process is carried out on multiple timescales. In our investigation we use this meaning of the term 'capability planning'.

Defense capability planning is usually based on the scenario method. Scenarios represent different frames for plausible futures [16]. Some of these frames are aligned with current world views and strategies, while others deviate from this alignment. Their aim is to focus the planner's mind on establishing future contexts and help make a case for the development and justification of strategies. They also assist the planner in defining intermediate goals along the path towards the future.

Common to all scenarios is that they deal with uncertainties and illustrate major issues a planner has to deal with. Schoemaker [16] sees the elements in the scenario set as elements that bound uncertainty. Thus, one can think of scenarios as the edges that bound a multi-dimensional sub-space of uncertainty.

In the defence and other literature, uncertainty falls into two classes: branching points and surprises/shocks linked to deep uncertainty. Davis and Kahan [7] define deep uncertainty as "materially important uncertainties that cannot be adequately treated as simple random processes and that cannot realistically be resolved at the time they come into play". One of the causes for deep uncertainty are sudden and unique incidents, or *wild cards*, "that can constitute turning points in the evolution of a certain trend" [14]. Hence, deep uncertainty itself is unpredictable which precludes it from being modeled in trend analysis, forecasting methods or by employing other probability-based techniques.

Stated differently, deep uncertainty is "the result of pragmatic limitations in our ability to express all that we know about complex adaptive systems and their associated policy problems" [3]. This form of uncertainty thus poses a significant challenge for the capability planner, and an important step towards tackling this challenge is to get rid of the desire of precise prediction and the goal of eliminating uncertainty through accurate forecasts [15] Eden [10] argues that the primary reason to account for uncertainty in strategic planning is that it turns planning for the future from a one-off event to an ongoing learning proposition.

With learning a key objective, the aim of techniques in support of the planning process shifts away from prediction accuracy and the requirement for a 'best' solution. Instead, providing the planner with tools to explore the space of possible futures and a means of generating information on future alternatives becomes important. Scenario planning, apart from being most defense organizations' tool of choice, provides such an environment for futures exploration. Here we define scenario planning as a continuous process for capturing dynamic causality in the eyes of uncertainty to identify alternatives and informative courses of actions that management can pursue, when faced with fast-moving changes.

While uncertainty – in particular discontinuities and limitations arising from deep uncertainty – has been and still is a challenge in scenario planning, complexity is the emerging one. Recent advances in the management literature show a vast trend in research on complexity [4][11]. However, it is rare to see complexity addressed in the scenario planning literature – with the exception of a few studies in defense [2][7][8][17][18].

There are two main reasons why capability planning is not only complicated but complex. Firstly, capability systems are composites of subsystems that have intricate and non-linear spatiotemporal interdependencies and often interact strongly with the environment in which they are embedded. This is particularly the case for defense capabilities. For instance, planning for a pre-emptive strike capability would involve the balancing of air, maritime and land capabilities; it would change a national defense strategy and hence affect current and future defense capabilities; and its realization would need to consider political, social and cultural sensitivities. Secondly, capability planners are not just 'passive recipients' of what the future might bring. They actively influence and shape the future context in which the capability will operate. In the aforementioned example, the introduction of a pre-emptive strike capability could actually spark and feed the very threat it wants to defeat; and the implications on the political and social environment, both domestic and international, are huge.

Some of the consequences of this complexity are that (1) phase transitions, self-organization and other phenomena of non-linear dynamics can occur, and thus simple cause-effect relationships do no longer govern the causality of the planning problem; (2) planning outputs may show large variations in response to small changes in inputs; and (3) capability system boundaries are hard or even impossible to define. Scenario planning therefore is in need of techniques that help unveil causality hidden in the non-linearities of the problem; explore the results of input changes; and interpret the consequences of planning actions. For this, we propose to complement the scenario technique with fast, frugal and exhaustive exploration tools that provide outputs of alternatives and summarize the benefits and drawbacks of such alternatives in a meaningful way.

In this paper, we introduce what we call Computational Scenario-based Capability Planning (CSP). CSP is *not* about computerizing the strategic planning process. CSP has two objectives: (1) expanding the horizon of scenario planning to manage the complexity arising from networking future operations; and (2) enhancing the scenario planning process with computer models that make use of the advances in Information and Communication Technology (ICT) and assist the planner and management in the exploration of the space of possibilities. In CSP, the first objective is accomplished by adopting principles from the study of complex adaptive systems [13]. The second objective is achieved through computer-in-the-loop scenario planning based upon novel interactive models of contexts and the computerization of parts of the scenario generation process. Because planning is a human-centric process, we use '*computer*-in-the-loop', i.e. the planners are fully integrated in the process and use the computational models as exploratory and decision aiding tools. Thus ICT 'guides' the planning process. This paper will only focus on the second objective by introducing the methodology and illustrating it in an abstract – but representative – case study.

## 2. Planning Problems are Not Optimization in Uncertain or Dynamic Environments Problems

Probably the highest gain organizations achieve from scenario planning is the use of scenarios for reshaping the mind of the decision maker towards perceiving uncertainty, current and future vulnerabilities, and risks that otherwise would not be perceived. This is to a certain degree captured in Donald Rumsfeld's famous saying that "Plans are nothing, planning is everything". Planning, including scenario-based planning, trains our minds to face the unknown.

The key differentiator between scenario planning and traditional prediction methods is the way uncertainty is handled. In traditional probabilistic approaches, scenarios can be seen as events with their associated probability distributions. This approach is successful in a number of domains such as describing the spread of diseases. In the strategic planning arena, however, the probabilistic view of the world has many limitations that make it unfit.

The first problem is the massive amount of data that need to be collected, maintained and updated to manage these probability distributions. The second problem is its inability to account for emotions, feelings, and the complex human behavior. The socio-technical interface is critical in many, if not all, strategic security and defence planning exercises. The third problem is it can handle one type of uncertainty only: branching points - the planned uncertainty that we understand well and can map onto a set of probability distributions [6]. Shocks and surprises can't be accounted for. Consequently, the approach suffers from an inadequacy to capture discontinuities in systems. Some would argue that September 11, the collapse of the Soviet Union, the rise of China and the fall of the Roman empire caused discontinuities in history. These discontinuities are tipping points in the life-cycle of the system and cannot be modeled by the probabilistic and analytical schools. The fourth problem is that the use of historical data to generate the probability distributions – assuming the continuity of trends - is inadequate beyond short-term planning.

As such, thinking of the planning problem as an optimization in an uncertain and dynamic environment can be misleading. Firstly, the objective of optimization is to find the best global or local solution. In planning, however, there is no such thing as the best solution for the problem. The objective of planning is to unfold the uncertainty in future environments and identify strategies to harness this uncertainty. Secondly, optimization in dynamic environments assumes that there is a trend underneath the change; otherwise there is no advantage from adaptation. Deep uncertainty represents a discontinuity in the strategic space, and as such, can change the parameters as well as the structure of a problem. This makes the concept of adaptive optimization obsolete. Thirdly, in an uncertain environment, multiple evaluations of the same solution configuration would result in different values of the objective function. The variation in these values is normally attributed to noise. Therefore, most approaches rely on calculating the average (normalized) fitness of a solution assuming that a solution is implemented an infinite number of times to overcome the noise. When evaluating a solution in scenario planning, variations in the evaluations are attributed to variations in the scenario space (i.e. the space of possibilities). Therefore, taking an average is not recommended since every evaluation represents a plausible future rather than noise. Taking the worst case is also misleading as it can result in an over-estimate of the required resources.

In this paper, the reader will develop an appreciation of the complexity of the planning problem and will find out that optimization techniques for uncertain and dynamic environments can be used if the aforementioned pitfalls are avoided. However, optimization techniques, alone, are insufficient to handle the complexity of a planning problem.

## 3. Resource Planning under Time Constraints

The real world capability planning problems can take multiple forms. This paper is part of a larger study on land mobility requirements and the design of future land mobility capability. We present here part of the methodology to answer the question of how many field vehicles an army needs in 2025. Our paper focuses on the structured part of the methodology with the actual process for answering the previous question being far more complex and outside the scope of this paper.

The methodology relies on the existence of a simulation environment for wargaming and an optimization environment (we call the Solver) for the identification of the minimum amount of resources required to achieve a certain level of efficiency. In fact, the high fidelity solver we built can be used to answer three questions: what is the minimum number of vehicles needed to fulfill the requirements of a scenario which – after an abstraction step – can be parameterized in terms of a certain number of tasks? Given an existing fleet, what is the efficiency of this fleet – measured, for instance, by the total number of fulfilled tasks – in a certain scenario? Given an existing fleet, how many vehicles of each type need to be acquired to meet the requirements of a given scenario?

The Solver is of high fidelity and contains a large number of specific military constraints (see [1] and [2] for more details). In this paper, however, neither the space nor the depth required for the analysis would make the use of the high-fidelity Solver feasible or practical. Therefore, we demonstrate our CSP methodology using an abstract version of the original problem. The abstract problem captures the complexity of the original problem while making it feasible to demonstrate a complex methodology. It is an amalgamation of the resource planning problem (i.e. find the amount of resources to meet a certain demand) and the resource-constrained project management problem (i.e. schedule tasks under resource constraints). We call the problem "resource planning under time constraints" (RPTC). A formal definition of RPTC is as follows:

Given

- a vector of vehicle types $V=(V_1, \ldots, V_n)$
- a cost vector, $C=(C_1, \ldots, C_n)$ representing the cost of each vehicle type $T_i$,
- a vector of resource types $R=(R_1, \ldots, R_m)$
- a matrix $M=\{M_{ij}\}$, with $M_{ij}$ representing the amount of resources of type j that one unit of a vehicle type $V_i$ can transport,
- a list of tasks $T=\{T_{lj}\}$, $l=1\ldots k_j$, with $k_j$ representing the number of tasks associated with resource type j; and $T_{lj}=(D_{lj},ES_{lj},MD_{lj},Q_{lj})$, with $D_{lj}$ representing the duration of the task labeled with lj, $ES_{lj}$ representing the earliest start time in hours and minutes, $MD_{lj}$ representing the maximum delay in minutes allowed, and $Q_{lj}$ representing the number of units of resource j required.
- a cost vector, $B=(B_1, \ldots, B_m)$ representing the cost that one unit of a resource of type $R_j$ is not fulfilled in one of the tasks associated with that resource type.

Find

- a vector $X=(X_1,\ldots,X_n)$, where $X_i$ is the number of vehicles needed of type $V_i$, such that all tasks are fulfilled

by performing the following optimizations

- Minimize the total cost of vehicles, C.X
- Minimize the variance var(X) of the vehicle vector.

The first objective is the result of tight cost constraints in the land mobility capability project. The second objective captures in a simple way the requirement for robust fleets, i.e. fleets which are stable against unexpected task variations and most likely to adjust to large, unplanned changes in task profiles. Experience shows that this characteristic can be approximated by 'balancing' the fleet, i.e. by minimizing the difference between the number of vehicles of different types. While equating robustness with var(X) minimization is inaccurate, it is sufficient for illustrating our CSP methodology.

The two objectives are not necessarily in conflict. However, they conflict with each other in many of the scenarios we generated, especially when there is a demand for many different vehicle types. Therefore, it is safe to assume that the problem is a multi-objective optimization problem. In this paper, we use NSGA-II [9] to optimize this problem.

The vector X is evaluated by a simple heuristic. This heuristic assigns the most costly unfulfilled task to the largest available vehicle that can fulfill the task, locks the vehicle for the duration of the task, reduces the task requirements by the amount moved by the vehicle, then loops until all tasks are fulfilled. If at the end some tasks remain unfulfilled, the total cost of an unfulfilled task is calculated, gets multiplied by a penalty term and added to objective functions.

While occasionally this simple heuristic may determine sub-optimal vehicle vectors X, it would be disadvantageous to replace it with a more complicated one. As indicated in the Introduction, the aim of planning techniques is to be explorative while prediction accuracy is of minor importance. Elaborate heuristics can (marginally) improve the accuracy of a single RPTC solution; however, this comes at the expense of increased processing time. In the CSP the solution of a single RPTC problem is only one point in a set of alternatives generated during an exhaustive scenario exploration phase that may require solving the RPTC problem thousands or millions of times. Thus exploration speed becomes essential, and simple, fast heuristics outperform potentially more precise but slower optimization techniques. In addition, the methodology we present in the next section is robust against the occasional sub-optimality of individual RPTC solutions because these solutions get aggregated and the advantage of having an exact optimal solution for each RPTC gets reduced or may even disappear. Nevertheless, this does not mean that we should not strive to find optima. If sub-optimality is frequent then the aggregation over-estimates the vehicles required in the land mobility capability. The point we want to make is that there is a compromise. An investigation of the algorithm used in the high-fidelity solver [2] revealed that some simple heuristics are so powerful that they can generate very close to optimal solutions in time and resource constrained scheduling problems.

| Figure 1. The Computational Planning Model |
|---|

**Stage 1: Scenario Generation**

Step 1. Use Creative Thinking Methods to identify deep uncertainties and design different futures scenario structures.

Step 2. Build a database of future scenario structures; call it SDB. Initialize SDB.counter to the number of scenarios in SDB

Step 3. Select a different scenario from SDB, subtract 1 from SDB.counter.

**Stage 2: RPTC Sampling**

Step 4. Parameterize the scenario.

Step 5. Use human-based or computer-based simulations to generate the required list of tasks for that scenario.

Step 6. Formulate the RPTC problem

Step 7. Add the problem to SDB(i).PDB(j)

Step 8. Solve the RPTC problem

Step 9. Extract the Non-dominated set where all tasks in the scenario are fulfilled and add it to the list of NDS, and let NDS(l).cost be the cost of acquiring the vehicles corresponding to solution l.

Step 10. If stopping criteria is not met, go to Step 4

Step 11. If SDB.counter >0, go to Step 3.

Step 12. Evaluate every non-dominated solution l in NDS on every problem in SDB(i).PDB(j) for all i and j, by measuring the total cost of unfulfilled tasks to generate an overall score NDS(l).score(j) for each scenario j

**Stage 3: Recommendations**

Step 13. Use k-centroid clustering to cluster the list of NDS such that the maximum distance between any two weighted solutions in a cluster should not exceed a threshold $\theta_1$.

Step 14. For each cluster $\Psi_\alpha$, $\alpha=1…\Theta$, where $\Theta$ is the total number of clusters found in Step 13, find $\Omega_\alpha$ representing the ceil on the number of vehicles in each cluster $\alpha$. , evaluate the ceil on all problem instances for all scenarios to calculate the score function.

Step 15. Use k-centroid to cluster $\Omega_\alpha$ such that the maximum distance between any two weighted solutions in a cluster does not exceed a threshold $\theta_2$.

Step 16. If either the instances fall in a single cluster or all clusters have one element, go to Step 17, else go to Step 14.

Step 17. Use the hierarchy of clusters generated from Steps 13-16 to build the capability evolution network.

Step 18. Use the scores assigned on each solution to calculate a score on each scenario for each node in the network.

## 4. Proposed Computational Planning Model

The proposed methodology has three main stages. Stage 1 is responsible for generating scenarios. Stage 2 is responsible for sampling instances of the RPTC problem and solving it. Stage 3 is responsible for grouping recommendations across all scenarios and problem instances to develop a suitable path from the current fleet mix to future fleet mix alternatives. Stages 1 and 2 are connected through a feedback loop, where scenarios can be generated based on observations made during Stage 1. However, in this paper, this feedback is not considered because of the complexity it imposes on the overall process. The three stages can be assumed to be independent. Figure 1 depicts the steps underlying each stage. In the next section, a case study is used to explain each step to reduce repetition and maximize clarity.

## 5. Case study

In this section, we present a hypothetical case study, to elaborate on the CSP methodology and demonstrate realistic concepts.

## 5.1 Planning Purpose

Assume that:

- today is 1$^{st}$ of January 2000;

- there is a budget of $0.5b available to buy additional vehicles in the next 5 years;

- there are six different types of vehicles: two armored vehicles ($V_1$,$V_2$), two specialized engineering vehicles for construction ($V_3$,$V_4$), and two class-B vehicles for supplying ammunitions, food, water, etc. ($V_5$,$V_6$);

- the cost of each vehicle is $0.2m, $0.4m, $0.2m, $0.4m, $0.5m and $0.8m respectively;

- there are seven different resources types ($R_1$,…,$R_7$), each vehicle can only work on one task at a time (i.e. a vehicle cannot do two different types of tasks simultaneously);

- the cost of not fulfilling a task for each resource type is $100, $200, $50, $200,$ 20, $200, and $100, respectively;

- $V_1$ can deliver up to two and four units of resource $R_1$ and $R_2$, respectively; $V_2$ can deliver up to three, six and two units of resource $R_1$, $R_2$ and $R_3$, respectively; $V_3$ can deliver up to five, ten and eight units of resource $R_5$, $R_6$ and $R_7$, respectively; $V_4$ can deliver up to eight, twelve and 14 units of resource $R_5$, $R_6$ and $R_7$, respectively; $V_5$ can deliver up to four and three units of resource $R_3$ and $R_4$, respectively; $V_6$ can deliver up to ten units of resource $R_4$ only;

- The question is how many vehicles should we order today based on our anticipated fleet mix in 2025.

In the following subsections, we will demonstrate the methodology using a hypothetical – but meaningful – example.

## 5.2 Stage 1: Scenario Generation

Step 1. <u>Use Creative Thinking Methods to identify deep uncertainties and design different futures scenario structures.</u>

Scenario planning is chosen as the approach to answer the previous question. A committee of experts undertook a number of brainstorming sessions and the Field Anomaly Relaxation (FAR) technique [5] was used to identify the basic factors underpinning future operations. FAR identified three factors for shaping the future, these are: Fire superiority (F), economic, social and political negotiation abilities (E), and logistics (L). For each factor, three levels of High (denoted as 1), Medium (denoted as 2) and Low (denoted as 3) prevalence were defined. A possible future can be seen as the combination of the three factors at different levels. For example, $F_3E_1L_1$ may represent a future where the military powers of different nations are equal and conflicts are resolved in negotiations, or a future where humanitarian operations are important. Another example is $F_1E_3L_1$ representing a future where offensive operations are dominant. The FAR analysis revealed three plausible futures: $F_3E_1L_1$, $F_1E_3L_1$, and $F_3E_3L_1$. These represent the three futures to be investigated.

The sole purpose of this step is to capture deep uncertainty. Each of the three scenarios mentioned above create discontinuity in the strategy space, where one plan that is suitable for one scenario is not fit for the other. In reality, deep uncertainty is far more complex to capture than this simple hypothetical example, and we are developing more rigorous approaches to capture it. However, the advantage of the three scenarios we have lies in the discontinuity they will create in our strategy space.

Step 2. Build a database of future scenario structures; call it SDB. Initialize SDB.counter to the number of scenarios in SDB

Each of these futures is normally scripted in a narrative that defines the backbone structure of a scenario. Each narrative represents a generic sequence of how the dynamics of a scenario may unfold. One can think of each of these as a movie, where characters are given general guidelines of their roles and behavior without being told exactly what to do. In step 5, the characters play the movie and the dynamics (i.e. the story) unfold as the movie is played.

Step 3. Select a different scenario from SDB, subtract 1 from SDB.counter.

This is an iterative step, where each scenario is selected in turn. We work on all scenarios simultaneously for ease of presentation.

## 5.3 Stage 2: RPTC Sampling

In this stage, the scenarios are parameterized, simulated to generate the tasks, the RPTC problems are formulated and solved, and the solutions are grouped and evaluated on all scenarios.

Step 4. Parameterize the scenario.

The parameterization of each scenario is the process of defining the functional requirements for a scenario to be simulated. For example, in scenario $F_3E_1L_1$, we need to define the red (enemy) and blue (friendly) forces, the context in which negotiation will take place, the different behaviors red may exhibit, etc. At the end of this stage, the complete simulation setup is well-defined so that the scenario can be simulated in the next step.

Step 5. Use human-based or computer-based simulations to generate the required list of tasks for that scenario.

The military is famous for conducting live experiments, where an entire operation can be simulated by a group of humans. However, these human-based simulations or experimentations are very expensive. Instead, computer-based simulations can be a cheaper alternative. However, computer-based simulations are normally criticized for being pre-scripted and there is no room for cognitive and behavioral aspects to be represented efficiently. Recently, a type of multi-agent systems known as agent-based distillations [12][17][18] has shown potential in overcoming these limitations. Any of these agent-based distillation systems can be used to simulate a scenario, although we prefer the Warfare Intelligent

| Scenarios | Resources | Tasks Characteristics | | | |
|---|---|---|---|---|---|
| | | $k_j$ | $D_{lj}$ | $MD_{lj}$ | $Q_{lj}$ |
| $F_3E_1L_1$ | $R_1$&$R_2$ | High U(200,300) | Short N(5,1) | Critical U(0,2) | Low N(10,2) |
| | $R_3$&$R_4$ | High N(100,20) | Short N(5,1) | Critical U(0,2) | Medium N(20,4) |
| | $R_5$,$R_6$&$R_7$ | Low N(5,1) | Short N(5,1) | Critical U(0,2) | Low N(5000,100) |
| $F_1E_3L_1$ | $R_1$&$R_2$ | Low U(10,20) | Short N(5,1) | Not Critical U(0,60) | Low N(10,2) |
| | $R_3$&$R_4$ | Medium U(20,40) | Medium N(10,2) | Not Critical U(0,50) | Medium N(20,4) |
| | $R_5$,$R_6$&$R_7$ | High N(500,50) | Long N(5,1) | Not Critical U(0,40) | Low N(3000,200) |
| $F_3E_3L_1$ | $R_1$&$R_2$ | Low N(10,2) | M N(10,2) | Critical U(0,4) | Medium N(50,10) |
| | $R_3$&$R_4$ | Medium U(10,30) | Medium N(10,2) | Less Critical U(10,20) | High N(100,20) |
| | $R_5$,$R_6$&$R_7$ | Low N(5,1) | Long N(5,1) | Less Critical U(10,20) | Low N(5000,100) |

System for Dynamic Optimization of Missions (WISDOM), which we have developed for modeling combat operations. WISDOM has many advantages including its ability to reason on group level behaviors and to analyze in real-time many measures including social network analysis and combat measures.

Running WISDOM many times with different random seeds, discretizing the time into bins to generate the different tasks, collating these bins and fitting probability distributions to find out the probability distribution of each task type would generate the previous table. In the table, $k_j$ defines the observed frequency of tasks (High, Medium and Low); $D_{lj}$ defines the observed duration (Long, Medium and Short); $MD_{lj}$ defines the observed time criticality of tasks measured by the maximum delay allowed for a task (Critical, Less critical and Not critical); $Q_{lj}$ represents the observed demand quantity of the task (Large, Medium and Small); and the letters "N" and "U" in the table denote the fitted Normal and Uniform distributions respectively.

Step 6.  Formulate the RPTC problem

The probability distributions, along with their parameters, in the table can sample many instances of the RPTC. We sample 1000 instances for each scenario.

Step 7.  Add the problem to SDB(i).PDB(j)

Each instance of the RPTC that corresponds to each scenario will be added to the database SDB(i).PDB(j) with PDB initiated at the generation of the first scenario instance . The following figures show the frequency diagram for parameter $k_j$ summed over all resource types. Each figure represents one of the three scenarios. The differences in the task distributions among scenarios are indications of the discontinuity in the strategic space that will be reflected in the final set of options.

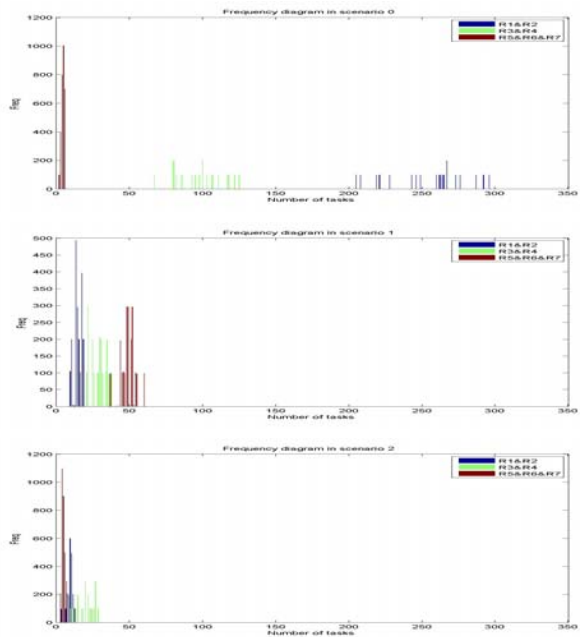

Step 8.  Solve the RPTC problem

Each problem instance for each scenario is solved using NSGA-II and the heuristic mentioned in Section 2. The population size is set to 50, and the maximum number of generations is set to 100. The default parameters for NSGA-II are used.

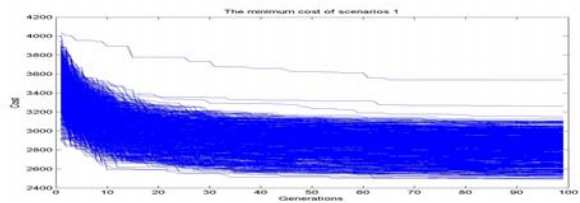

In the previous figure, the x-axis denotes generation identification time, and the y-axis the non-dominated solution with the minimum total cost in 1000 instances of the second scenario. The clear gap between the top curve in the figure and almost all other curves is a reflection of the uncertainty in that scenario. This entails that most instances can be solved with a fleet costing $2.4b to $3.3b, while one of the instances requires at least $3.5b.

Step 9.  Extract the non-dominated set (NDS) where all tasks in the scenario are fulfilled and add it to the list of NDS, and let NDS(l).cost be the cost of acquiring the vehicles corresponding to solution l.

The non-dominated set for each problem instance and for all scenarios is extracted. The following figure shows a scatter diagram of all non-dominated solutions generated for scenario 2. The gap shown in the previous figure is still clear in the different curves for non-dominated solutions.

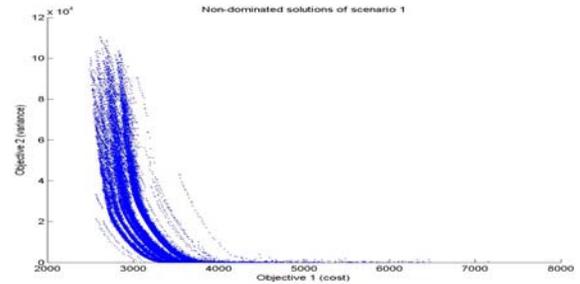

If we look at a frequency diagram of the total cost for acquiring vehicles for each of the three scenarios (shown in the next three figures), one can see the effect of discontinuity in the scenario space on the distribution of solutions that are fit for different scenarios. In essence, each scenario can be seen as a different problem, so it is natural to have different solution set. The key in strategic planning is to find a path over time to solve all these problems. As such, these problems are different but not un-related.

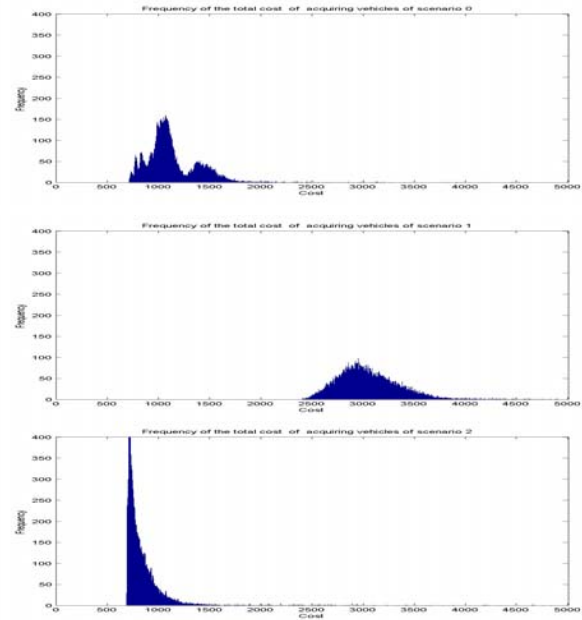

Step 10.  If stopping criteria is not met, go to Step 4

This is an iterative step. The stopping criterion is to solve all problem instances for the scenario.

Step 11.  If SDB.counter >0, go to Step 3.

This is an iterative step. The stopping criterion is to solve all scenarios.

Step 12. Evaluate every non-dominated solution i in NDS on every problem in SDB(j).PDB(k) for all j and k, by measuring the total cost of unfulfilled tasks to generate an overall score NDS(i).score(j) for each scenario j

In this step, every non-dominated solution that got generated for each problem instance is evaluated on all scenarios. A score is given for each solution as follows

$$NDS(i).score(j) = \sum_k \frac{CUT_k^j}{NOT_k^j}$$

where $CUT_k^j$ is the cost for unfulfilled tasks for solution i when measured on problem instance k in scenario j, and $NOT_k^j$ is the number of tasks in problem instance k in scenario j.

All scores are then normalized for each scenario independently by dividing each score by the maximum score obtained in a scenario. The score function represents the risk of adopting a non-dominated solution as measured by the level of vulnerability resulting from not fulfilling some tasks in different scenarios.

## 5.4 Stage 3: Recommendations

Step 13. Use k-centroid clustering to cluster the list of NDS such that the maximum distance between any two weighted solutions in a cluster should not exceed a threshold $\theta_1$.

We use k-centroid clustering carried out on the solution/genotype. The clustering operation is preformed under the constraint of the weighted solution; that is, each number of vehicles is replaced by the number of vehicles multiplied by the cost of a vehicle. Therefore, the distance between two weighted solutions represent the Euclidean distance between the cost vector of two possible fleet mixes. The threshold $\theta_1$ – taken in this study to be $0.5b - represents the difference in total costs for which two solutions are seen to be invariant or similar in terms of their budget.

Step 14. For each cluster $\Psi_\alpha$, $\alpha=1…\Theta$, where $\Theta$ is the total number of clusters found in Step 13, find $\Omega_\alpha$ representing the ceil on the number of vehicles in each cluster $\alpha$, evaluate the ceil on all problem instances for all scenarios to calculate the score function.

The ceil for each cluster is calculated. It represents the minimum number of vehicles from each type required in a fleet that can perform all problem instances that generated the solutions in that cluster. Each ceil is evaluated on all problem instances and the score function is calculated for each scenario.

Step 15. Use k-centroid to cluster $\Omega_\alpha$ such that the maximum distance between any two weighted solutions in a cluster does not exceed a threshold $\theta_2$.

This is a recursive step, where the ceils are also clustered, in a similar way to the original solutions, and new ceils are calculated. The idea of this step is to build a hierarchy of clusters which aggregate the capabilities. $\theta_2$ is taken to be $0.5b.

Step 16. If either the instances fall in a single cluster or all clusters have one element, go to Step 17, else go to Step 14.

This is a conditional step to decide on whether or not to continue the clustering. In our case study, we generated approximately 150,000 non-dominated solutions from all runs. After removing duplicates (the same solution generated by different runs), this was reduced to 139,769 non-dominated solutions. The clustering algorithm generated 341 clusters. After taking the ceil of each cluster and applying the clustering algorithm on the ceil, 174 clusters were generated. The algorithm became stable with 162 clusters.

Discussing this large network or visualizing it in this paper was not possible. Therefore, we decided to filter the data by only selecting the solution with the least cost in each non-dominated set. This generated 3,000 solutions (3 scenarios x 1000 instances per scenario). Clustering these solutions resulted in 29 clusters. By taking the ceil for each cluster and clustering these ceils, 18 clusters were generated and the algorithm became stable.

Step 17. Use the hierarchy of clusters generated from Steps 13-16 to build the capability evolution network.

The hierarchical clusters generated from Steps 13-16 are used to generate the capability evolution network on next page. Notice that the origin of the network represents our present land mobility capability. Each transition in the network is bounded by $\theta_2$, which means that each transition costs no more than $\theta_2$.

Step 18. Use the scores assigned on each solution to calculate a score on each scenario for each node in the network.

A capability evolution network shows the possible transitions from one capability set to a larger capability set. Each transition will be from a low cost capability set to a higher cost capability set, as well as, from a high score value (high risk) to a lower score value (low risk). Each node of the capability evolution network is then parameterized with the scores of the configurations on that node and each arc is parameterized with the difference in cost between the configurations are each end. We visualize the capability evolution network in the next page. The network clearly shows the complexity of the problem.

## 6. Conclusion

In this paper, we motivated the need for new fast and frugal tools that facilitate the exploration of capability planning problems and thus offer learning opportunities to the strategic planners. We formulated the principles of the computational scenario-based capability planning (CSP) methodology which combines the scenario technique with modern ICT methods to allow for (1) the generation of vast sets of plausible futures from a few high-level scenario templates; (2) the exhaustive exploration of the space of scenario instantiations to unravel the complex causality hidden in such problems; (3) the formulation of conflicting objectives at different planning levels; (4) the interpretation of the potentially enormous set of generated capability solutions; and (5) the interaction between humans and machines through feedback (and feedforward) loops. At the core of the methodology, an evolutionary multi-objective optimization algorithm is used to generate capability options.

We applied a simplified version of the CSP methodology to a hypothetical case study of a land mobility capability planning process and showed how abstraction and parameterization of scenario templates can be used to generate thousands of scenario instantiations, and how conflicting planning objectives create vast amounts of non-dominant capability options. Clustering methods were employed to aid in the solution analysis.

The application of the CSP methodology to this case study highlights that it is a powerful approach for deciding on future capabilities. This paper is merely a first step towards the full CSP methodology and its application to real-world planning problems. A number of further developments are needed, in particular in the areas of scenario abstraction and solution interpretation. This is the focus of future work, in which we aim to improve the methodology and make it more practical by adopting the methodology to effect-based planning.

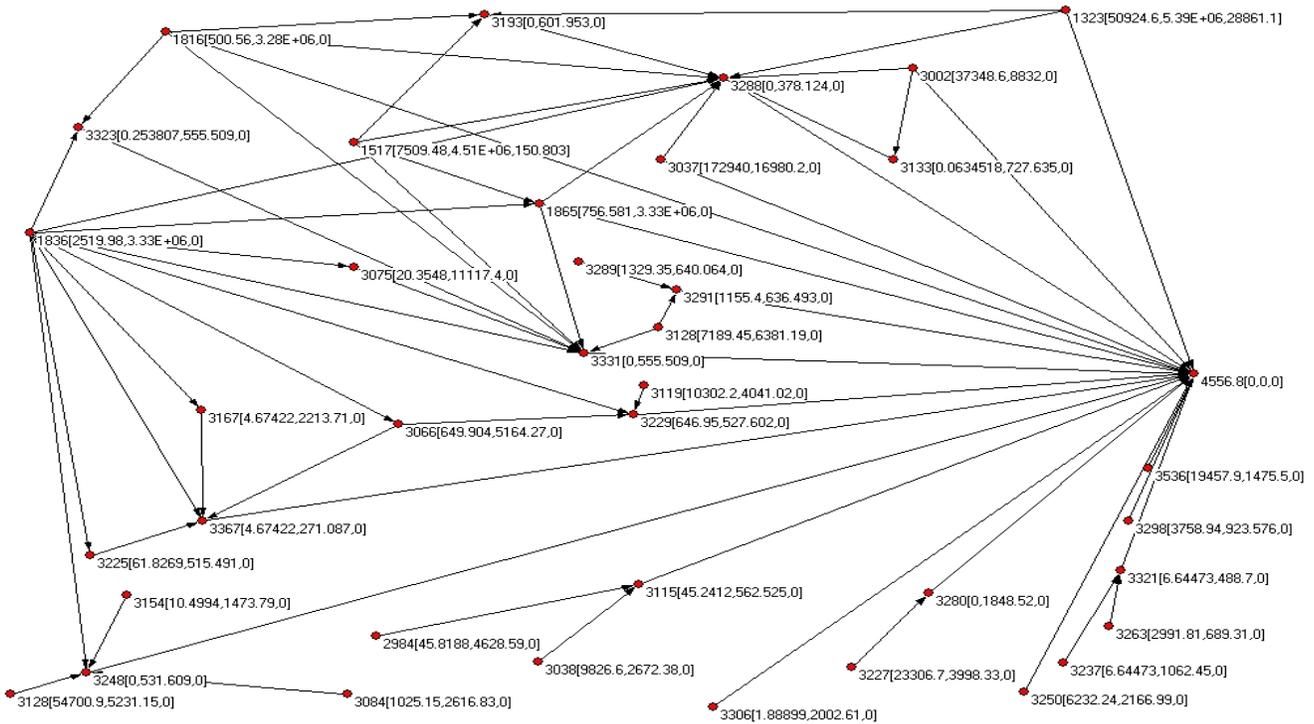

Technology Organisation, North Atlantic Treaty Organization (NATO): NEUILLY-SUR-SEINE CEDEX, France.

[20] Radford K.J., Strategic Planning: An Analytical Approach. 1980: Reston Publishing Company.